\documentclass[letterpaper]{article} 
\usepackage{aaai24}  
\usepackage{times}  
\usepackage{helvet}  
\usepackage{courier}  
\usepackage[hyphens]{url}  
\usepackage{graphicx} 
\usepackage{natbib}  
\usepackage{caption} 
\usepackage{algorithm}
\usepackage{algorithmic}

\usepackage{newfloat}
\usepackage{listings}
\DeclareCaptionStyle{ruled}{labelfont=normalfont,labelsep=colon,strut=off} 
\floatstyle{ruled}
\newfloat{listing}{tb}{lst}{}
\floatname{listing}{Listing}
\usepackage{graphicx}
\usepackage{amsmath}
\usepackage{booktabs}
\usepackage{array}
\usepackage{amsfonts}

\title{Enhancing Robotic Manipulation with AI Feedback from Multimodal Large Language Models}
\author{
    Jinyi Liu\textsuperscript{\rm 1}\equalcontrib, Yifu Yuan\textsuperscript{\rm 1}\equalcontrib, Jianye Hao\textsuperscript{\rm 1}, Fei Ni\textsuperscript{\rm 1}, Lingzhi Fu\textsuperscript{\rm 1}, Yibin Chen\textsuperscript{\rm 1}, Yan Zheng\textsuperscript{\rm 1}  \\
}
\affiliations{
    \textsuperscript{\rm 1}{College of Intelligence and Computing, Tianjin University}
    \\
    \{jyliu, yuanyf, jianye.hao, fei\_ni, lingzhifu, yibin\_chen, yanzheng\}@tju.edu.cn
}

\usepackage{bibentry}

\begin{document}

\maketitle

\begin{abstract}
Recently, there has been considerable attention towards leveraging large language models (LLMs) to enhance decision-making processes. However, aligning the natural language text instructions generated by LLMs with the vectorized operations required for execution presents a significant challenge, often necessitating task-specific details. To circumvent the need for such task-specific granularity, inspired by preference-based policy learning approaches, we investigate the utilization of multimodal LLMs to provide automated preference feedback solely from image inputs to guide decision-making. In this study, we train a multimodal LLM, termed CriticGPT, capable of understanding trajectory videos in robot manipulation tasks, serving as a critic to offer analysis and preference feedback. Subsequently, we validate the effectiveness of preference labels generated by CriticGPT from a reward modeling perspective. Experimental evaluation of the algorithm's preference accuracy demonstrates its effective generalization ability to new tasks. Furthermore, performance on Meta-World tasks reveals that CriticGPT's reward model efficiently guides policy learning, surpassing rewards based on state-of-the-art pre-trained representation models.
\end{abstract}

\section{Introduction}

In recent years, the emergence of pre-trained large language models (LLM) has marked a significant milestone in the advancement of artificial intelligence (AI), enhancing the performance and generalization capability of AI models in handling text-based tasks. Notable exemplars include ChatGPT~\cite{chatgpt} and LLaMa~\cite{Llama2}. Furthermore, the integration of multimodal understanding, encompassing video, images, audio, and text, has led to the proposal of multimodal LLM (MLLM), significantly advancing the perceptual comprehension of multimodal information by LLM. Noteworthy models comprise GPT-4v, LLaVA~\cite{llava}, and EmbodiedGPT~\cite{embodiedGPT}. Building upon the robust generalization capacity of MLLM in perceiving the visual information of the physical world, a salient concern emerges: \textit{how to leverage MLLM's comprehension of visual information to enhance decision-making process in visual tasks}.

Recent endeavors have sought to leverage LLMs to drive decision-making processes, providing behavior instruction tailored to the current context~\cite{ChatGPTforRobotics, doasican, InnerMonologue}. However, these initiatives face significant obstacles, such as aligning natural language suggestions with the practical execution of actions~\cite{yu2023language}, as well as aligning textual depictions of current state information with actual state. To mitigate the alignment issues between textual instructions and actual action vectors, some literature have proposed to use a finer-grained and more direct numerical information - reward signal - to bridge the gap between LLMs' output and the policy learning process~\cite{yu2023language, Self-Refined}, with LLMs generating rewards to facilitate policy learning. However, these approaches heavily rely on task-specific details, such as the heights of robot joints, thereby constraining their adaptability to novel tasks. Without the need for prior knowledge of task-specific details, our goal is to generate reward signals for novel visual tasks solely by leveraging the advanced comprehension and generalization capabilities of MLLMs towards visual information.

Traditionally, experts design manual reward values in a progressive manner to ensure that a fully completed task state yields a higher reward than its incomplete counterparts. Take the task of opening a door as an example: the reward value for the state of touching the door handle should surpass that of not touching it, yet remain less than the reward for both touching the handle and successfully opening the door. We propose to apply this reasoning to MLLM, enabling it to comprehend task-specific visual information and discern which states are more conducive to task completion, thereby allocating higher reward signals.

In this paper, we introduce a multimodal LLM, CriticGPT, which serves as a \textbf{critic} capable of analyzing and comparing different video trajectories for the same task and providing preference feedback. Subsequently, leveraging these autonomously generated preference feedback, we can then fit a reward model capable of generating dense reward signals to facilitate policy learning.
This approach shares similarities with Reinforcement Learning from Human Feedback (RLHF)~\cite{bai-2204-05862}. 
However, the focus of this study is on leveraging feedback signals from CriticGPT, to achieve automated, high-quality preference feedback, thereby reducing human resource costs. To the best of our knowledge, this work represents the inaugural effort to align Reinforcement Learning with AI Feedback (RLAIF)~\cite{ConstitutionalAI, RLAIF} in visual decision-making scenarios, using robot manipulation as a primary example.

One significant challenge arises from the limited support provided by existing open-source multimodal LLM, which are mostly fine-tuned on general image datasets. To address this, we first collect a large and diverse set of video data for robot manipulation tasks and then fine-tune LLaVA-1.5~\cite{llava1.5}, a well-performing open-source MLLM, to obtain CriticGPT. The fine-tuning process endows MLLM with the capability to comprehend video inputs and evaluate trajectories of robotic manipulation tasks. The results show that CriticGPT demonstrates efficient understanding for robot videos in robot manipulation tasks, achieving a preference accuracy of over 95\% on our test dataset.

To illustrate the advantages of using CriticGPT to drive decision-making as a critic, we conduct experiments on the Meta-World benchmark~\cite{metaworld}. We compare the performance of learning algorithms under dense rewards provided by the environment, reward models (RM) fitted based on CriticGPT feedback, and visual representation rewards (LIV~\cite{ma2023liv}) under the DrQ-v2~\cite{YaratsFLP22DRQV2}. The results show that in the majority of environments, learning under our RM is more effective. Notably, CriticGPT's reward model demonstrates robust performance on novel tasks not included in the fine-tuning dataset, showcasing the generalization capability of our approach.

This research makes three main contributions. Firstly, we propose CriticGPT as a powerful tool for providing effective analysis and preference feedback from video data, serving as the first instance of RLAIF in robot manipulation. Secondly, experimental results demonstrate that CriticGPT feedback can effectively generalize to new robot manipulation tasks, achieving learning efficiencies that match or surpass those under sparse reward or reward based on pre-trained presentation model. Thirdly, we construct a large-scale multimodal decision dataset containing a wealth of trajectory videos and corresponding analyses, applicable for model performance evaluation and fine-tuning.

\section{Related Work}

This section provides an overview of the related work involved in this research.

\paragraph{Multimodal Large Language Models.}
The advancement of Large Language Models~(LLM) has seen a significant surge following the emergence of ChatGPT~\cite{chatgpt}, and subsequent advancements including GPT 4~\cite{GPT4} and PaLM~\cite{palm2}. Subsequently, an increasing number of open-source models have been continually introduced, aiming to achieve the performance of top-tier closed-source models, such as the pure-text LLaMA~\cite{Llama2}, Vicuna~\cite{vicuna2023}, and the multimodal LLM LLaVA~\cite{llava}, LLaMA-Adapterv2~\cite{LLaMA-Adapterv2} and others~\cite{BLIP-2, minigpt4, Otter, InstructBLIP, Video-LLaMA, mllmsurvey}.

\paragraph{Pre-trained Vision-Language Representation Model} 
Pre-trained vision-language representation models achieve the concurrent extraction of image and text features by jointly learning from extensive unlabeled datasets. Prominent examples include CLIP~\cite{clip}. In our study, we employ CLIP as the image encoder.

Pre-trained representation models are also used for reward generation, via measuring the distance between representations, exemplified by VIP~\cite{vipMaSJBK023} anb LIV~\cite{ma2023liv}. This aligns with the foundational premise of our research, wherein, in the absence of detailed structured information, a generalizable method is employed to learn the reward function from image information. Notably, the rewards generated by these methods hinge on representation, whereas our approach is grounded in the MLLM's comprehension of semantic information in image trajectories.

\paragraph{Large Language Models for Control Tasks.}
Building on the understanding of decision-making scenarios by LLMs, advanced work has begun to focus on how to use LLMs to empower the decision-making process. 
Currently, most works focus on how to use LLM to understand natural language instructions related to the task, and then provide suggestions on actions to be executed~\cite{ChatGPTforRobotics,doasican,InnerMonologue,SocraticModels,CodeasPolicies,yu2023language}. These methods design different ways to align LLM's output (usually natural language) with downstream policy execution, such as affordance function~\cite{doasican}, intrinsic reward~\cite{Guidingpretraining}, foundation model~\cite{vima}, pre-defined interface function~\cite{ChatGPTforRobotics}, and so on. In addition, some literatures~\cite{yu2023language, Self-Refined} focus on reward generation based on the output of LLMs, given pre-defined APIs or sufficient prior knowledge about the task (such as joint height). In contrast, our work does not seek any task-specific prior knowledge when facing new tasks, but rather focuses on the MLLM's understanding and evaluation capabilities of visual trajectories, using AI preferences as the link to guide policy learning.
\begin{figure*}[h]
  \centering
  \includegraphics[width=0.92\textwidth]{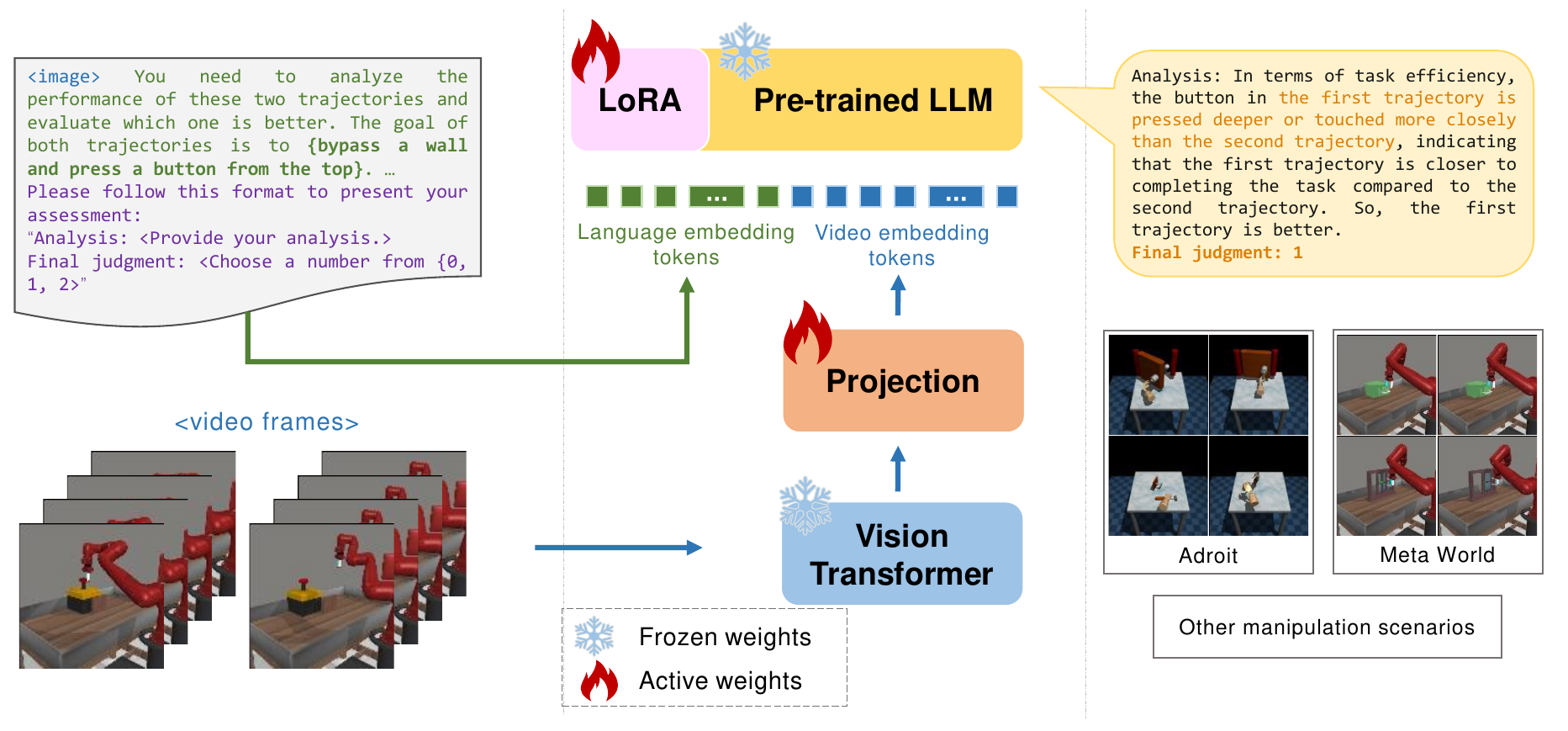}
  \caption{The architecture of CriticGPT, along with concise input-output examples. CriticGPT accommodates video input and, based on natural language instructions, generates corresponding responses.}
  \label{fig:finetuning}
\end{figure*}

\paragraph{Reinforcement Learning from Preference Feedback}
Reinforcement learning is typically guided by reward signals provided by experts. However, in some scenarios, designing reward signals by experts can be challenging. Thus, various automated reward modeling approaches have been proposed to learning from preference feedback~\cite{bai-2204-05862, pebble, ChristianoLBMLA17}, including Reinforcement Learning from Human Feedback~(RLHF)~\cite{bai-2204-05862}. RLHF models reward preferences and aligns them with the policy training process. In the training process of the latest closed-source large models like ChatGPT~\cite{chatgpt} and PaLM2~\cite{palm2}, RLHF have been employed and are considered key to performance enhancement. Nevertheless, the heavy human involvement in the RLHF training process has been a significant limiting factor, inspiring researchers to explore AI's ability to provide feedback in lieu of human intervention, known as Reinforcement Learning from AI Feedback~(RLAIF)~\cite{ConstitutionalAI}. A recent work~\cite{RLAIF} demonstrates that AI feedback can achieve performance comparable to human feedback on text-related tasks. In this paper, we focus on revealing the feedback capability of Multimodal LLM on robot manipulation tasks, addressing the challenge of high human feedback costs.

\section{CriticGPT: A Multimodal LLM as a Critic}

In this paper, we aim to facilitate policy learning from raw image data in novel tasks without detailed prior knowledge, by harnessing the robust visual understanding capabilities of MLLM. However, existing MLLM like LLaVA~\cite{llava1.5} lack comprehension of trajectory videos for robotic manipulation tasks. To address this gap, we fine-tune LLaVA to introduce a novel instruction-following assistant, termed CriticGPT. Compared to LLaVA, CriticGPT has expanded capabilities: it can comprehend video inputs rather than single image only, and can understand robotic manipulation videos, providing insightful analysis and judgment as a critic. 
Furthermore, we discuss how to leverage the feedback provided by CriticGPT more effectively to facilitate reinforcement learning policy learning by fitting the reward model.
This section elaborates on the technical intricacies of these processes.

\subsection{Visual Instruction-Following Dataset for Robot Manipulation}
To align the existing model with trajectory images and videos in robot manipulation tasks, we collect instruction-following dataset for fine-tuning, employing a fully automated script to collect a substantial amount of instruction-following data on classic control tasks, i.e., Meta-World~\cite{metaworld}.

For two distinct trajectory videos $\textbf{X}_\mathrm{v}$, we generate inquiries $\textbf{X}_\mathrm{instruct}$ with a sampled question $\textbf{X}_\mathrm{q}$ under that task as well as response instruction $\textbf{X}_\mathrm{i}$, and corresponding natural language analysis of visual trajectories $\textbf{X}_\mathrm{a}$ and evaluation results $\textbf{X}_\mathrm{e}$. We organize these information into a single round of instruction-following dialogue, which includes: 
\begin{equation}
\begin{aligned}
\texttt{Human:} \quad & \textbf{X}_\mathrm{instruct} = \text{Random}  \left\{
\begin{array}{ll}
\left[\textbf{X}_\mathrm{v}, \textbf{X}_\mathrm{q}, \textbf{X}_\mathrm{i}\right]  \\[1pt]
\left[\textbf{X}_\mathrm{q}, \textbf{X}_\mathrm{i}, \textbf{X}_\mathrm{v}\right] 
\end{array}
\right. ,\\
\texttt{Assistant:} \quad & \left[\textbf{X}_\mathrm{a}, \textbf{X}_\mathrm{e} \right] .
\notag
\end{aligned}
\end{equation}
Among these, $\textbf{X}_\mathrm{q}$ represents the questions specific to a particular task. For example, in the case of the ``button-press-wall" task, the description is like: ``This video merges two robot motion trajectory segments. You need to analyze the performance of these two trajectories and evaluate which one is better. The goal of both trajectories is to \textit{bypass a wall and press a button}. ". Note that only the task-specific description varies for different tasks, and in our collected dataset, we uniformly use the descriptions provided in the original paper~\cite{metaworld} for automated generation. $\textbf{X}_\mathrm{i}$ represents the instructions for the output, including macro guidance on how to judge which trajectory is better and guidance on the output format. This ensures a reduction in the possibility of irrelevant or nonsensical responses and facilitates the evaluation of model accuracy. 

The analysis of visual trajectories, $\textbf{X}_\mathrm{a}$, primarily derives from information that can be automatically extracted from the task environment. This information includes details such as task completion status, reward, distance to the goal, and more. Using natural language templates and rule-based scripts, we can link together this god's-eye view information to create descriptive analyses of video trajectories. These analyses can be seen as guiding the inference to obtain the final evaluative result (i.e., preference) about the quality of the trajectories, $\textbf{X}_\mathrm{e}$, in a Chain of Thought (COT) manner.

\begin{figure*}[t]
  \centering
  \includegraphics[width=0.76\textwidth]{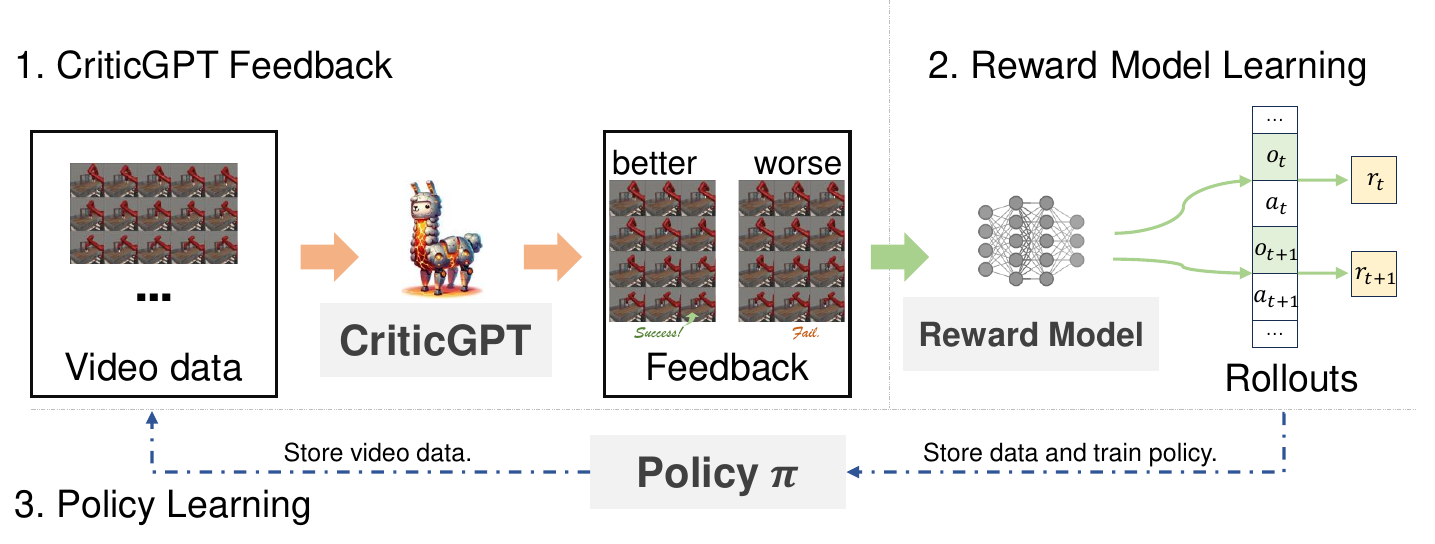}
  \caption{An overview of using automated feedback labels generated by CriticGPT to facilitate policy learning. Without introducing ambiguity, details regarding collecting video data and the transition buffer are omitted.}
  \label{fig:fig1}
\end{figure*}
\subsection{Adapting MLLM to Robot Manipulation Scenarios as a Critic}

Based on the collected visual instruction-following data, we fine-tune LLaVA-1.5 to obtain CriticGPT, making it to be able to understand video input and analyse robot manipulation trajectory videos. 
Our adopted network architecture closely resembles that of LLaVA, featuring a pre-trained vision encoder and a Projection layer to bridge the vision encoder with the LLM, facilitating improved adaptation of vision embeddings within the LLM. 

Figure~\ref{fig:finetuning} depicts the architecture of CriticGPT. During inference, video frames are input into the Vision Transformer to obtain video embeddings, which are then passed through a Projection layer to produce video embedding tokens, aligning them more effectively with language tokens. The video tokens from the Projection layer are concatenated with the language tokens derived from the natural language instruction, feeding into the pre-trained large language model. Finally, the large language model outputs a detailed analysis and evaluation of the visual trajectories as a critic.

During training, given the collected visual instruction-following data, we maintain the frozen parameters in the vision encoder, utilizing its embeddings directly as video frame representations. To conserve training cost and prevent the degradation of LLM's generalization due to domain-specific databases, we also freeze the parameters of the LLM and fine-tune it using LoRA~\cite{lora}. We also perform fine-tuning on the Projection layer to ensure better alignment between video embeddings and language tokens.

After fine-tuning, we obtain an MLLM model, CriticGPT, proficient in understanding robot video datasets. Beyond its ability to comprehend video data, a pivotal feature of CriticGPT lies in its capacity to analyze and provide preference feedback on robotic manipulation trajectory videos, functioning as a critic. The Critic possesses more directly information, thereby empowering various downstream tasks. Subsequently, we delineate the process of using automatically generated preference feedback with CriticGPT to guide control policy learning.

\subsection{Policy Training}

Figure~\ref{fig:fig1} illustrates the complete process of how CriticGPT aids in policy learning. Initially, video data is provided, and CriticGPT autonomously generates analytical and evaluative outcomes, including preference labels. Subsequently, based on the collected label data, a reward model is trained to align with CriticGPT's judgments, following a basic preference-based learning framework to learn a reward function $\hat{r_\psi}$ from these preferences~\cite{WilsonFT12,ChristianoLBMLA17,pebble}. Finally, during the interaction between the reinforcement learning algorithm and the environment, this reward model is utilized to generate reward signals for different image observations, thereby guiding policy learning.

For two trajectory segments $\sigma^0$ and $\sigma^1$, let's denote the preference as $y$. Then, a preference predictor using the reward function $\hat{r_\psi}$ can be modeled as follows:
\begin{equation}
P_\psi[\sigma^1\succ\sigma^0]=\frac{\exp\sum_t\widehat{r}_\psi(\mathbf{s}_t^1,\mathbf{a}_t^1)}{\sum_{i\in\{0,1\}}\exp\sum_t\widehat{r}_\psi(\mathbf{s}_t^i,\mathbf{a}_t^i)},
\end{equation}
where $\sigma^1\succ\sigma^0$ represent that trajectory segment $1$ is preferable to another one. We use the same approach in PEBBLE~\cite{pebble} to train this dense reward function, parameterised by $\psi$:
\begin{equation}
\begin{aligned}
    \mathcal{L}^{\text{Reward}} = - \mathbb{E}_{(\sigma^0,\sigma^1,y)\sim\mathcal{D}}\Big[&y(0)\log P_\psi[\sigma^0\succ\sigma^1] \\&+y(1)\log P_{\psi}[\sigma^1\succ\sigma^0]\Big] .
\end{aligned}
\end{equation}

\begin{table*}[h]
\caption{Comparison of judgement accuracy on the collected dataset.}
\small
\label{tab:acc}
\renewcommand{\arraystretch}{1.0}
\centering
\begin{tabular}{>{\raggedright}m{2.6cm}>{\raggedright}m{0.9cm}>{\raggedleft}m{0.9cm}>{\raggedleft}m{0.9cm}>{\raggedleft}m{1.5cm}>{\raggedleft}m{1.8cm}>{\raggedleft}m{1.6cm}} 
    \toprule
    Model \textbackslash\ Dataset & Train. & Test & Hard Test & plate-slide-side & button-press-topdown-wall & button-press (hard) \tabularnewline
   \midrule
LLaVA-1.5              & 0.439                      & 0.412                     & 0.452                          & 0.478                                & 0.444                                         & 0.550                                 \tabularnewline 
CriticGPT (12.5\%) & 0.916                      & 0.943                     & 0.545                          & 0.889                                & 0.942                                         & 0.631                                  \tabularnewline
CriticGPT (25\%)   & 0.921                      & 0.970                     & 0.542                          & 0.882                                & 0.855                                         & 0.590                                 \tabularnewline
CriticGPT (50\%)   & 0.925                      & \textbf{0.990}            & 0.499                          & 0.981                                & \textbf{1.000}                                & 0.606                                  \tabularnewline
CriticGPT (bs=64)  & 0.939                      & 0.972                     & 0.545                          & 0.967                                & 0.892                                         & 0.565                                  \tabularnewline
\midrule
CriticGPT          & \textbf{0.944}             & \textbf{0.999}            & \textbf{0.706}                 & \textbf{1.000}                       & \textbf{1.000}                                & \textbf{0.746}      \tabularnewline \bottomrule                  
\end{tabular}
\end{table*}

After obtaining a reward model aligned with CriticGPT's preferences for visual trajectories, we use this reward model $\hat{r_\psi}$ to generate reward. We employ DrQ-v2~\cite{YaratsFLP22DRQV2}, an off-policy algorithm known for efficient sampling, for training image-input policies. The training data is a collection of tuples $(s_t, a_t, r_t, d_t, s_{t+1})$, where $s_t, a_t, d_t, s_{t+1}$ are obtained through interactions with the environment, and $r_t$ is generated by the reward model $\hat{r_\psi}$.

\section{Experiment}
We conduct experiments and focus on two aspects: firstly, how does CriticGPT perform? Secondly, is the preference feedback based on CriticGPT efficient for downstream policy learning?

\begin{figure*}[h]
    \centering
    \includegraphics[width=0.95\textwidth]{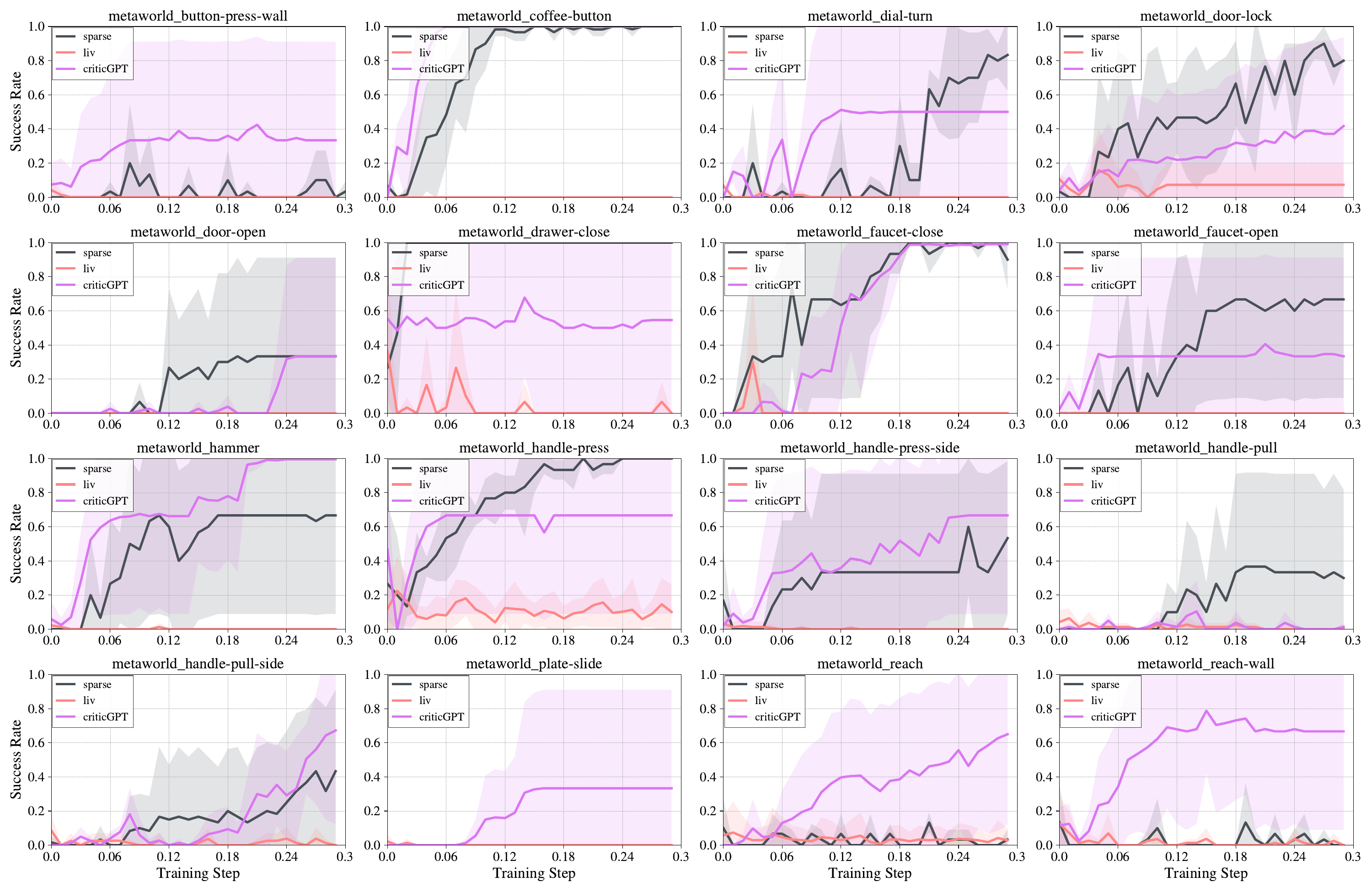}  
    \caption{Results of DrQ-v2 with different reward on the Meta-World benchmark.}
    \label{fig:eval}
\end{figure*}
\begin{figure*}[h]
    \centering
    \includegraphics[width=0.95 \textwidth]{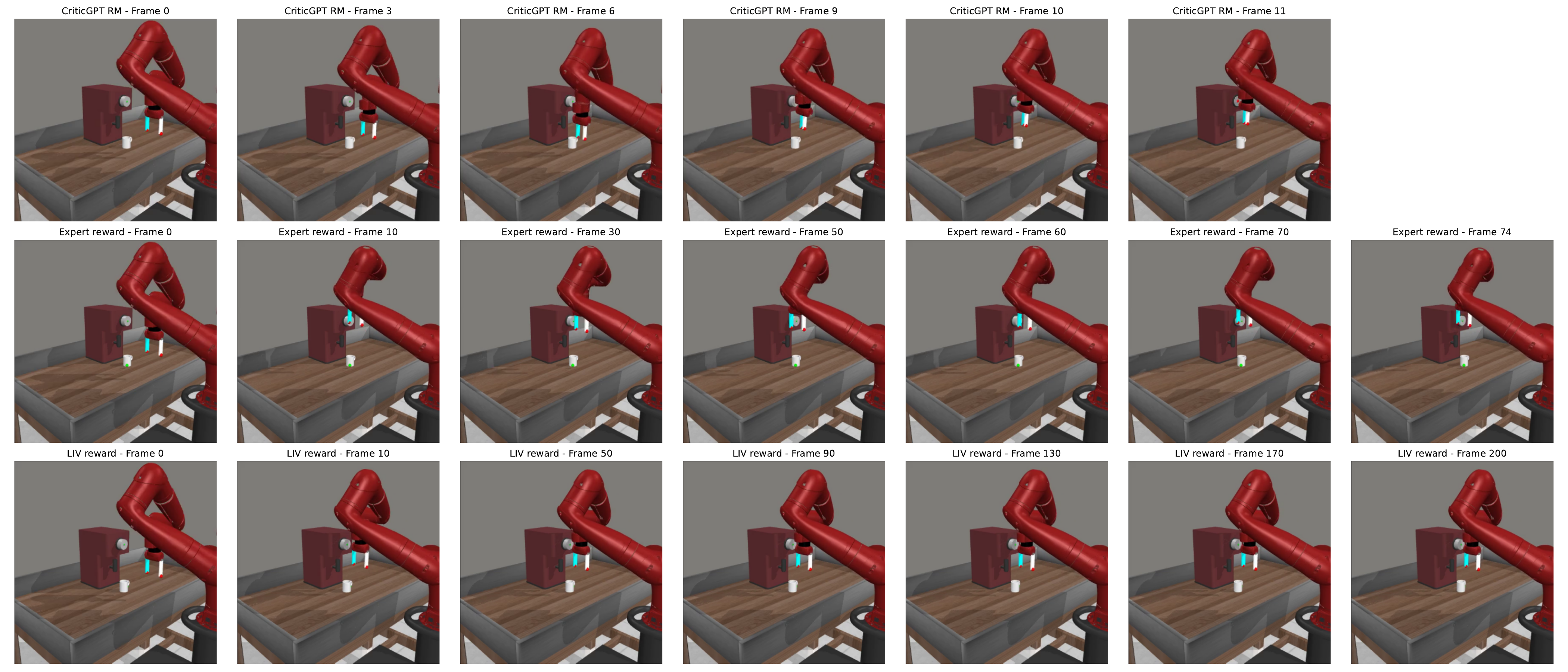}  
    \caption{Illustrating differences in behavioral performance under various rewards using the coffee-button task as an example. Trajectories achieving success or near-success around 40k training steps are selected, and their initial, final, and intermediate frames are visualized.}
    \label{fig:video}
\end{figure*}

\subsection{Performance of Fine-Tuned CriticGPT}
\paragraph{Collecting Dataset}
We initially collect instruction-following data in the Meta-World benchmark~\cite{metaworld}. The reason for choosing this benchmark is primarily because it provides a wealth of interpretable information, such as task completion status and distances to the target, which can be directly obtained. This information aids in the automated generation of explanations for state trajectory images, facilitating reasoning about the superiority or inferiority of two trajectories in a COT-like thought process. We use SAC~\cite{SAC} models trained at different stages to generate trajectories of varying quality. These trajectories are randomly paired to form query pairs, and then, based on detailed information such as rewards and task completion status, we automatically generate COT analyses and judgments through Python scripts. We collect 1.5K question-answer pairs for each task, resulting in a training set of 24K samples with 16 tasks, and a testing set of 7.5K samples with 5 tasks.
\paragraph{Performance of CriticGPT}

To assess the efficacy of CriticGPT as a critic and elucidate the impact of various factors (training duration, batch size) on fine-tuning effectiveness, we compare the baseline model, LLaVA-1.5, with multiple intermediate models of CriticGPT. To evaluate model effectiveness, we compare the preference labels output by CriticGPT with the ground truth labels, calculating the accuracy. We exclude instructions leading to a 0 output (indicating similar trajectory quality), prompting the model to provide binary judgments. Table~\ref{tab:acc} presents the accuracy of these models on the dataset. The model (x\%) means different training period, and CriticGPT~(bs=64) represents the model trained under the batch size as 64, instead of 32 as default. 

We report accuracy not only on the training and test sets but also on additional datasets: the Hard Test dataset represents a collection of trajectory pairs from 5 tasks with comparable performance trajectory pairs, posing a challenge to the model's judgment. The datasets plate-slide-side and button-press-topdown-wall represent a task in the training set and test set, respectively, demonstrating the model's maximum performance on individual task datasets. button-press~(hard) represents a set of trajectory pairs with comparable performance on this task.

Our findings reveal that the general-purpose MLLM (LLaVA) falls considerably short in comprehending trajectory pairs, analyzing differences, and aligning to instructions, underscoring the need for fine-tuning MLLM in specific control scenarios. CriticGPT demonstrates exceptionally high accuracy under typical circumstances, highlighting MLLM as a potent tool for analyzing and reasoning about trajectory variations. In exceedingly challenging scenarios~(Hard datasets), CriticGPT performs slightly higher than random performance (due to binary judgments, random accuracy is 0.5). This is primarily attributed to the inherent difficulty in distinguishing similar data, as human annotators often assign 0 judgments in such cases. Nevertheless, improving evaluation accuracy for such data remains an imminent priority.

It is noteworthy that due to limitations in the diversity of collected data, LLaVA's accuracy approaches 100\% on many datasets. However, this does not imply reaching performance limits or overfitting. Instead, enhancing evaluation methods and datasets is an important consideration for future research.

\subsection{Efficiency of CriticGPT Preference Feedback for Policy Learning}
\begin{figure}[h]
    \centering
    \includegraphics[width=0.48 \textwidth]{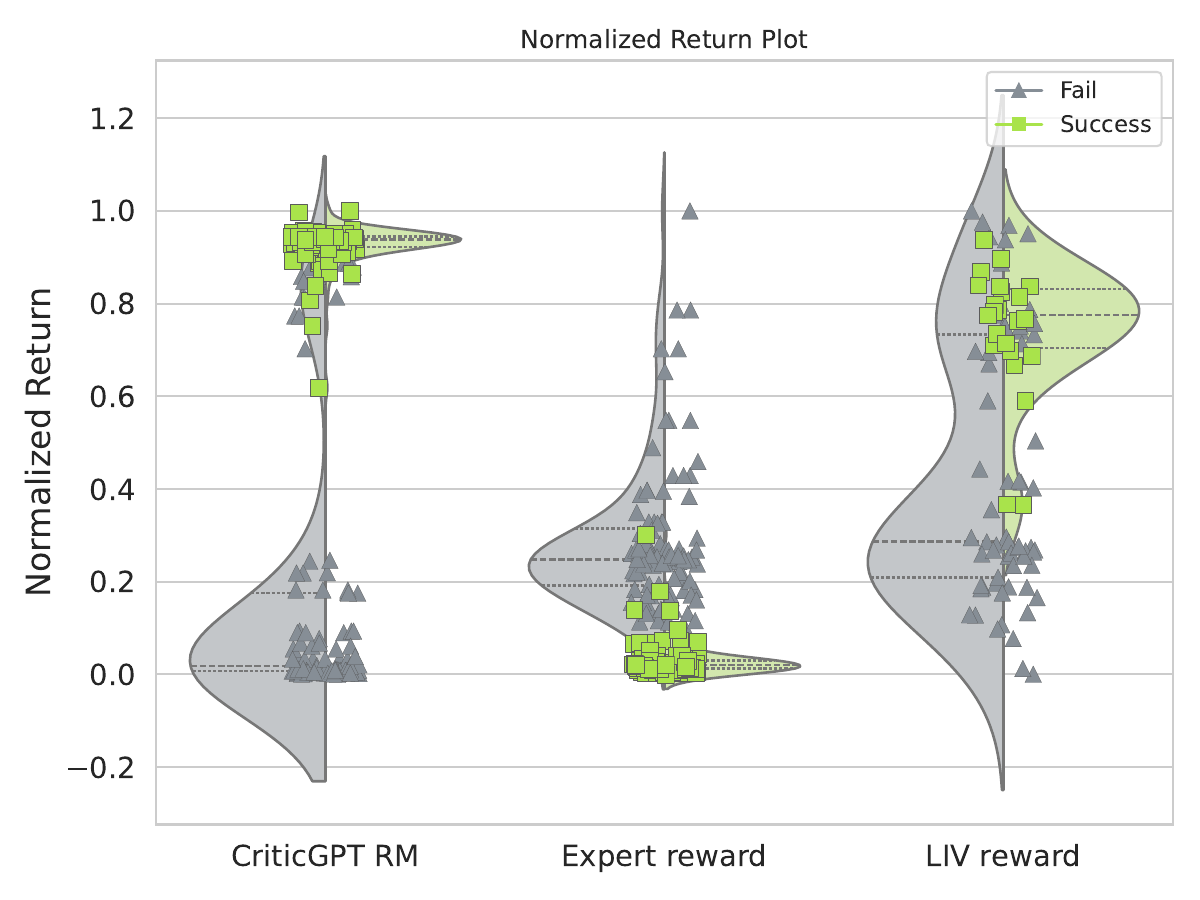}  
    \caption{Comparative analysis of the distribution differences in cumulative episode rewards obtained by different reward methods, with scattered points representing normalized cumulative reward values for different trajectories.}
    \label{fig:return}
\end{figure}

The preceding experiments verify that CriticGPT, compared to the general MLLM, LLaVA-1.5, exhibits more effective understanding of video inputs, especially robot manipulation trajectory videos, and generates reasonable preference feedback. To further illustrate the value of CriticGPT for downstream tasks, we evaluate the effectiveness of fitting a reward model based on CriticGPT preferences to guide policy learning.

The experimental results are presented in Figure~\ref{fig:eval}, where the curves and shaded areas represent the mean and standard error of success rates during 300K training steps across three seeds (set to 0, 1, 2). The baseline algorithms include the original sparse reward function indicating the completion or non-completion of a task, and the latest pre-trained representation reward model, LIV~\cite{ma2023liv}. The results indicate that the performance guided by CriticGPT RM is superior, underscoring the significant impact of CriticGPT RM on facilitating task completion. Some tasks, such as reach, button-press-wall, reach-wall, and coffee-button, are not included in the fine-tuning dataset. CriticGPT's superior performance on these tasks highlights its generalization and broad applicability to new tasks in the robot manipulation domain.

The results shown in Figure~\ref{fig:eval} exhibit several characteristics. Firstly, the performance of the LIV reward is poor. Our analysis reveals that the scale of rewards provided by LIV is small, and the rewards for different states are very close, making it difficult to differentiate between them. This issue may require further fine-tuning of the LIV model and enhancing its representation capabilities to address. Secondly, under the guidance of CriticGPT reward, the learning curve plateaus in the later stages instead of continuing to improve. This is one of the current limitations of the reward model training, indicating that the performance of the reward model collapses in the later stages and fails to accurately capture the preference order relationship. This is a crucial problem we are currently addressing, namely, how to enhance the robustness of reward model training.

Subsequently, we investigate why CriticGPT preference yield outstanding results in the derived RM. Firstly, Figure~\ref{fig:video} illustrates the differences in behavioral performance under three reward functions. It is evident that the policy guided by CriticGPT RM is more direct, swiftly pressing the button and completing the task. In contrast, the policy guided by the original expert reward takes longer to complete the task, and the behavior guided by the representation-based LIV reward is very close to task completion and persists in such near-completion state for an extended period (from the 10th frame to the 200th frame), but not success. These observations suggest that CriticGPT RM is more goal-oriented, while the expert reward introduces numerous small rewards guiding task completion, potentially slowing down the efficiency. The representation-based LIV reward may be hindered by insufficient training, making it challenging to distinguish between states close to completion and states indicating task completion, resulting in behaviors of prolonged near-completion states.

We also conduct a statistical analysis of the sum of rewards given by different reward functions for multiple trajectories and their relationship with task completion, as shown in Figure~\ref{fig:return}. It is evident that the expert reward assigns higher cumulative reward values to a significant number of failed trajectories due to its step-by-step guidance, inadvertently impeding policy learning. The LIV reward behaves more reasonably, attributing higher reward values to successfully completed trajectories; however, due to the instability in the representation learning process, LIV reward fails to demonstrate a clear advantage within the 150K training steps. CriticGPT RM's assigned rewards appear more reasonable, with a distinct margin in the distribution of return values between successful and failed trajectories. This observation explains the impressive performance of CriticGPT RM within the 300K training period.

\section{Conclusion}
We introduce CriticGPT, a multimodal LLM capable of analyzing and evaluating trajectory videos in robot manipulation tasks. Leveraging CriticGPT's feedback, we train a reward model and guide downstream policy learning, demonstrating the effectiveness of AI feedback based on CriticGPT. The capabilities of CriticGPT are expected to strengthen with the continual expansion of the dataset. We are committed to ongoing efforts with the anticipation that CriticGPT can empower a broader spectrum of visual robot tasks.

\bibliography{main}

\begin{thebibliography}{37}
\providecommand{\natexlab}[1]{#1}

\bibitem[{Anil et~al.(2023)Anil, Dai, Firat, Johnson, Lepikhin, Passos, Shakeri, Taropa, Bailey, Chen, Chu, Clark, Shafey, Huang, Meier{-}Hellstern, Mishra, Moreira, Omernick, Robinson, Ruder, Tay, Xiao, Xu, Zhang, {\'{A}}brego, Ahn, Austin, Barham, Botha, Bradbury, Brahma, Brooks, Catasta, Cheng, Cherry, Choquette{-}Choo, Chowdhery, Crepy, Dave, Dehghani, Dev, Devlin, D{\'{\i}}az, Du, Dyer, Feinberg, Feng, Fienber, Freitag, Garcia, Gehrmann, Gonzalez, and et~al.}]{palm2}
Anil, R.; Dai, A.~M.; Firat, O.; Johnson, M.; Lepikhin, D.; Passos, A.; Shakeri, S.; Taropa, E.; Bailey, P.; Chen, Z.; Chu, E.; Clark, J.~H.; Shafey, L.~E.; Huang, Y.; Meier{-}Hellstern, K.; Mishra, G.; Moreira, E.; Omernick, M.; Robinson, K.; Ruder, S.; Tay, Y.; Xiao, K.; Xu, Y.; Zhang, Y.; {\'{A}}brego, G.~H.; Ahn, J.; Austin, J.; Barham, P.; Botha, J.~A.; Bradbury, J.; Brahma, S.; Brooks, K.; Catasta, M.; Cheng, Y.; Cherry, C.; Choquette{-}Choo, C.~A.; Chowdhery, A.; Crepy, C.; Dave, S.; Dehghani, M.; Dev, S.; Devlin, J.; D{\'{\i}}az, M.; Du, N.; Dyer, E.; Feinberg, V.; Feng, F.; Fienber, V.; Freitag, M.; Garcia, X.; Gehrmann, S.; Gonzalez, L.; and et~al. 2023.
\newblock PaLM 2 Technical Report.
\newblock \emph{CoRR}, abs/2305.10403.

\bibitem[{Bai et~al.(2022{\natexlab{a}})Bai, Jones, Ndousse, Askell, Chen, DasSarma, Drain, Fort, Ganguli, Henighan, Joseph, Kadavath, Kernion, Conerly, Showk, Elhage, Hatfield{-}Dodds, Hernandez, Hume, Johnston, Kravec, Lovitt, Nanda, Olsson, Amodei, Brown, Clark, McCandlish, Olah, Mann, and Kaplan}]{bai-2204-05862}
Bai, Y.; Jones, A.; Ndousse, K.; Askell, A.; Chen, A.; DasSarma, N.; Drain, D.; Fort, S.; Ganguli, D.; Henighan, T.; Joseph, N.; Kadavath, S.; Kernion, J.; Conerly, T.; Showk, S.~E.; Elhage, N.; Hatfield{-}Dodds, Z.; Hernandez, D.; Hume, T.; Johnston, S.; Kravec, S.; Lovitt, L.; Nanda, N.; Olsson, C.; Amodei, D.; Brown, T.~B.; Clark, J.; McCandlish, S.; Olah, C.; Mann, B.; and Kaplan, J. 2022{\natexlab{a}}.
\newblock Training a Helpful and Harmless Assistant with Reinforcement Learning from Human Feedback.
\newblock \emph{CoRR}, abs/2204.05862.

\bibitem[{Bai et~al.(2022{\natexlab{b}})Bai, Kadavath, Kundu, Askell, Kernion, Jones, Chen, Goldie, Mirhoseini, McKinnon, Chen, Olsson, Olah, Hernandez, Drain, Ganguli, Li, Tran{-}Johnson, Perez, Kerr, Mueller, Ladish, Landau, Ndousse, Lukosiute, Lovitt, Sellitto, Elhage, Schiefer, Mercado, DasSarma, Lasenby, Larson, Ringer, Johnston, Kravec, Showk, Fort, Lanham, Telleen{-}Lawton, Conerly, Henighan, Hume, Bowman, Hatfield{-}Dodds, Mann, Amodei, Joseph, McCandlish, Brown, and Kaplan}]{ConstitutionalAI}
Bai, Y.; Kadavath, S.; Kundu, S.; Askell, A.; Kernion, J.; Jones, A.; Chen, A.; Goldie, A.; Mirhoseini, A.; McKinnon, C.; Chen, C.; Olsson, C.; Olah, C.; Hernandez, D.; Drain, D.; Ganguli, D.; Li, D.; Tran{-}Johnson, E.; Perez, E.; Kerr, J.; Mueller, J.; Ladish, J.; Landau, J.; Ndousse, K.; Lukosiute, K.; Lovitt, L.; Sellitto, M.; Elhage, N.; Schiefer, N.; Mercado, N.; DasSarma, N.; Lasenby, R.; Larson, R.; Ringer, S.; Johnston, S.; Kravec, S.; Showk, S.~E.; Fort, S.; Lanham, T.; Telleen{-}Lawton, T.; Conerly, T.; Henighan, T.; Hume, T.; Bowman, S.~R.; Hatfield{-}Dodds, Z.; Mann, B.; Amodei, D.; Joseph, N.; McCandlish, S.; Brown, T.; and Kaplan, J. 2022{\natexlab{b}}.
\newblock Constitutional {AI:} Harmlessness from {AI} Feedback.
\newblock \emph{CoRR}, abs/2212.08073.

\bibitem[{Chiang et~al.(2023)Chiang, Li, Lin, Sheng, Wu, Zhang, Zheng, Zhuang, Zhuang, Gonzalez, Stoica, and Xing}]{vicuna2023}
Chiang, W.-L.; Li, Z.; Lin, Z.; Sheng, Y.; Wu, Z.; Zhang, H.; Zheng, L.; Zhuang, S.; Zhuang, Y.; Gonzalez, J.~E.; Stoica, I.; and Xing, E.~P. 2023.
\newblock Vicuna: An Open-Source Chatbot Impressing GPT-4 with 90\%* ChatGPT Quality.

\bibitem[{Christiano et~al.(2017)Christiano, Leike, Brown, Martic, Legg, and Amodei}]{ChristianoLBMLA17}
Christiano, P.~F.; Leike, J.; Brown, T.~B.; Martic, M.; Legg, S.; and Amodei, D. 2017.
\newblock Deep Reinforcement Learning from Human Preferences.
\newblock In \emph{Advances in Neural Information Processing Systems 30: Annual Conference on Neural Information Processing Systems 2017, December 4-9, 2017, Long Beach, CA, {USA}}, 4299--4307.

\bibitem[{Dai et~al.(2023)Dai, Li, Li, Tiong, Zhao, Wang, Li, Fung, and Hoi}]{InstructBLIP}
Dai, W.; Li, J.; Li, D.; Tiong, A. M.~H.; Zhao, J.; Wang, W.; Li, B.; Fung, P.; and Hoi, S. C.~H. 2023.
\newblock InstructBLIP: Towards General-purpose Vision-Language Models with Instruction Tuning.
\newblock \emph{CoRR}, abs/2305.06500.

\bibitem[{Du et~al.(2023)Du, Watkins, Wang, Colas, Darrell, Abbeel, Gupta, and Andreas}]{Guidingpretraining}
Du, Y.; Watkins, O.; Wang, Z.; Colas, C.; Darrell, T.; Abbeel, P.; Gupta, A.; and Andreas, J. 2023.
\newblock Guiding Pretraining in Reinforcement Learning with Large Language.
\newblock In \emph{International Conference on Machine Learning, {ICML} 2023, 23-29 July 2023, Honolulu, Hawaii, {USA}}, volume 202 of \emph{Proceedings of Machine Learning Research}, 8657--8677. {PMLR}.

\bibitem[{Gao et~al.(2023)Gao, Han, Zhang, Lin, Geng, Zhou, Zhang, Lu, He, Yue, Li, and Qiao}]{LLaMA-Adapterv2}
Gao, P.; Han, J.; Zhang, R.; Lin, Z.; Geng, S.; Zhou, A.; Zhang, W.; Lu, P.; He, C.; Yue, X.; Li, H.; and Qiao, Y. 2023.
\newblock LLaMA-Adapter {V2:} Parameter-Efficient Visual Instruction Model.
\newblock \emph{CoRR}, abs/2304.15010.

\bibitem[{Haarnoja et~al.(2018)Haarnoja, Zhou, Abbeel, and Levine}]{SAC}
Haarnoja, T.; Zhou, A.; Abbeel, P.; and Levine, S. 2018.
\newblock Soft Actor-Critic: Off-Policy Maximum Entropy Deep Reinforcement Learning with a Stochastic Actor.
\newblock In \emph{Proceedings of the 35th International Conference on Machine Learning, {ICML} 2018, Stockholmsm{\"{a}}ssan, Stockholm, Sweden, July 10-15, 2018}, volume~80 of \emph{Proceedings of Machine Learning Research}, 1856--1865. {PMLR}.

\bibitem[{Hu et~al.(2022)Hu, Shen, Wallis, Allen{-}Zhu, Li, Wang, Wang, and Chen}]{lora}
Hu, E.~J.; Shen, Y.; Wallis, P.; Allen{-}Zhu, Z.; Li, Y.; Wang, S.; Wang, L.; and Chen, W. 2022.
\newblock LoRA: Low-Rank Adaptation of Large Language Models.
\newblock In \emph{The Tenth International Conference on Learning Representations, {ICLR} 2022, Virtual Event, April 25-29, 2022}. OpenReview.net.

\bibitem[{Huang et~al.(2022)Huang, Xia, Xiao, Chan, Liang, Florence, Zeng, Tompson, Mordatch, Chebotar, Sermanet, Jackson, Brown, Luu, Levine, Hausman, and Ichter}]{InnerMonologue}
Huang, W.; Xia, F.; Xiao, T.; Chan, H.; Liang, J.; Florence, P.; Zeng, A.; Tompson, J.; Mordatch, I.; Chebotar, Y.; Sermanet, P.; Jackson, T.; Brown, N.; Luu, L.; Levine, S.; Hausman, K.; and Ichter, B. 2022.
\newblock Inner Monologue: Embodied Reasoning through Planning with Language Models.
\newblock In \emph{Conference on Robot Learning, CoRL 2022, 14-18 December 2022, Auckland, New Zealand}, volume 205 of \emph{Proceedings of Machine Learning Research}, 1769--1782. {PMLR}.

\bibitem[{Ichter et~al.(2022)Ichter, Brohan, Chebotar, Finn, Hausman, Herzog, Ho, Ibarz, Irpan, Jang, Julian, Kalashnikov, Levine, Lu, Parada, Rao, Sermanet, Toshev, Vanhoucke, Xia, Xiao, Xu, Yan, Brown, Ahn, Cortes, Sievers, Tan, Xu, Reyes, Rettinghouse, Quiambao, Pastor, Luu, Lee, Kuang, Jesmonth, Joshi, Jeffrey, Ruano, Hsu, Gopalakrishnan, David, Zeng, and Fu}]{doasican}
Ichter, B.; Brohan, A.; Chebotar, Y.; Finn, C.; Hausman, K.; Herzog, A.; Ho, D.; Ibarz, J.; Irpan, A.; Jang, E.; Julian, R.; Kalashnikov, D.; Levine, S.; Lu, Y.; Parada, C.; Rao, K.; Sermanet, P.; Toshev, A.; Vanhoucke, V.; Xia, F.; Xiao, T.; Xu, P.; Yan, M.; Brown, N.; Ahn, M.; Cortes, O.; Sievers, N.; Tan, C.; Xu, S.; Reyes, D.; Rettinghouse, J.; Quiambao, J.; Pastor, P.; Luu, L.; Lee, K.; Kuang, Y.; Jesmonth, S.; Joshi, N.~J.; Jeffrey, K.; Ruano, R.~J.; Hsu, J.; Gopalakrishnan, K.; David, B.; Zeng, A.; and Fu, C.~K. 2022.
\newblock Do As {I} Can, Not As {I} Say: Grounding Language in Robotic Affordances.
\newblock In \emph{Conference on Robot Learning, CoRL 2022, 14-18 December 2022, Auckland, New Zealand}, volume 205 of \emph{Proceedings of Machine Learning Research}, 287--318. {PMLR}.

\bibitem[{Jiang et~al.(2022)Jiang, Gupta, Zhang, Wang, Dou, Chen, Fei{-}Fei, Anandkumar, Zhu, and Fan}]{vima}
Jiang, Y.; Gupta, A.; Zhang, Z.; Wang, G.; Dou, Y.; Chen, Y.; Fei{-}Fei, L.; Anandkumar, A.; Zhu, Y.; and Fan, L. 2022.
\newblock {VIMA:} General Robot Manipulation with Multimodal Prompts.
\newblock \emph{CoRR}, abs/2210.03094.

\bibitem[{Lee et~al.(2023)Lee, Phatale, Mansoor, Lu, Mesnard, Bishop, Carbune, and Rastogi}]{RLAIF}
Lee, H.; Phatale, S.; Mansoor, H.; Lu, K.; Mesnard, T.; Bishop, C.; Carbune, V.; and Rastogi, A. 2023.
\newblock {RLAIF:} Scaling Reinforcement Learning from Human Feedback with {AI} Feedback.
\newblock \emph{CoRR}, abs/2309.00267.

\bibitem[{Lee, Smith, and Abbeel(2021)}]{pebble}
Lee, K.; Smith, L.~M.; and Abbeel, P. 2021.
\newblock {PEBBLE:} Feedback-Efficient Interactive Reinforcement Learning via Relabeling Experience and Unsupervised Pre-training.
\newblock In Meila, M.; and Zhang, T., eds., \emph{Proceedings of the 38th International Conference on Machine Learning, {ICML} 2021, 18-24 July 2021, Virtual Event}, volume 139 of \emph{Proceedings of Machine Learning Research}, 6152--6163. {PMLR}.

\bibitem[{Li et~al.(2023{\natexlab{a}})Li, Zhang, Chen, Wang, Yang, and Liu}]{Otter}
Li, B.; Zhang, Y.; Chen, L.; Wang, J.; Yang, J.; and Liu, Z. 2023{\natexlab{a}}.
\newblock Otter: {A} Multi-Modal Model with In-Context Instruction Tuning.
\newblock \emph{CoRR}, abs/2305.03726.

\bibitem[{Li et~al.(2023{\natexlab{b}})Li, Li, Savarese, and Hoi}]{BLIP-2}
Li, J.; Li, D.; Savarese, S.; and Hoi, S. C.~H. 2023{\natexlab{b}}.
\newblock {BLIP-2:} Bootstrapping Language-Image Pre-training with Frozen Image Encoders and Large Language Models.
\newblock In \emph{International Conference on Machine Learning, {ICML} 2023, 23-29 July 2023, Honolulu, Hawaii, {USA}}, volume 202 of \emph{Proceedings of Machine Learning Research}, 19730--19742. {PMLR}.

\bibitem[{Liang et~al.(2023)Liang, Huang, Xia, Xu, Hausman, Ichter, Florence, and Zeng}]{CodeasPolicies}
Liang, J.; Huang, W.; Xia, F.; Xu, P.; Hausman, K.; Ichter, B.; Florence, P.; and Zeng, A. 2023.
\newblock Code as Policies: Language Model Programs for Embodied Control.
\newblock In \emph{{IEEE} International Conference on Robotics and Automation, {ICRA} 2023, London, UK, May 29 - June 2, 2023}, 9493--9500. {IEEE}.

\bibitem[{Liu et~al.(2023{\natexlab{a}})Liu, Li, Li, and Lee}]{llava1.5}
Liu, H.; Li, C.; Li, Y.; and Lee, Y.~J. 2023{\natexlab{a}}.
\newblock Improved Baselines with Visual Instruction Tuning.

\bibitem[{Liu et~al.(2023{\natexlab{b}})Liu, Li, Wu, and Lee}]{llava}
Liu, H.; Li, C.; Wu, Q.; and Lee, Y.~J. 2023{\natexlab{b}}.
\newblock Visual Instruction Tuning.
\newblock \emph{CoRR}, abs/2304.08485.

\bibitem[{Ma et~al.(2023{\natexlab{a}})Ma, Kumar, Zhang, Bastani, and Jayaraman}]{ma2023liv}
Ma, Y.~J.; Kumar, V.; Zhang, A.; Bastani, O.; and Jayaraman, D. 2023{\natexlab{a}}.
\newblock {LIV:} Language-Image Representations and Rewards for Robotic Control.
\newblock In Krause, A.; Brunskill, E.; Cho, K.; Engelhardt, B.; Sabato, S.; and Scarlett, J., eds., \emph{International Conference on Machine Learning, {ICML} 2023, 23-29 July 2023, Honolulu, Hawaii, {USA}}, volume 202 of \emph{Proceedings of Machine Learning Research}, 23301--23320. {PMLR}.

\bibitem[{Ma et~al.(2023{\natexlab{b}})Ma, Sodhani, Jayaraman, Bastani, Kumar, and Zhang}]{vipMaSJBK023}
Ma, Y.~J.; Sodhani, S.; Jayaraman, D.; Bastani, O.; Kumar, V.; and Zhang, A. 2023{\natexlab{b}}.
\newblock {VIP:} Towards Universal Visual Reward and Representation via Value-Implicit Pre-Training.
\newblock In \emph{The Eleventh International Conference on Learning Representations, {ICLR} 2023, Kigali, Rwanda, May 1-5, 2023}. OpenReview.net.

\bibitem[{Mu et~al.(2023)Mu, Zhang, Hu, Wang, Ding, Jin, Wang, Dai, Qiao, and Luo}]{embodiedGPT}
Mu, Y.; Zhang, Q.; Hu, M.; Wang, W.; Ding, M.; Jin, J.; Wang, B.; Dai, J.; Qiao, Y.; and Luo, P. 2023.
\newblock EmbodiedGPT: Vision-Language Pre-Training via Embodied Chain of Thought.
\newblock \emph{CoRR}, abs/2305.15021.

\bibitem[{{OpenAI}(2022)}]{chatgpt}
{OpenAI}. 2022.
\newblock {ChatGPT}.

\bibitem[{OpenAI(2023)}]{GPT4}
OpenAI. 2023.
\newblock {GPT-4} Technical Report.
\newblock \emph{CoRR}, abs/2303.08774.

\bibitem[{Radford et~al.(2021)Radford, Kim, Hallacy, Ramesh, Goh, Agarwal, Sastry, Askell, Mishkin, Clark, Krueger, and Sutskever}]{clip}
Radford, A.; Kim, J.~W.; Hallacy, C.; Ramesh, A.; Goh, G.; Agarwal, S.; Sastry, G.; Askell, A.; Mishkin, P.; Clark, J.; Krueger, G.; and Sutskever, I. 2021.
\newblock Learning Transferable Visual Models From Natural Language Supervision.
\newblock In \emph{Proceedings of the 38th International Conference on Machine Learning, {ICML} 2021, 18-24 July 2021, Virtual Event}, volume 139 of \emph{Proceedings of Machine Learning Research}, 8748--8763. {PMLR}.

\bibitem[{Song et~al.(2023)Song, Zhou, Liu, Fang, Shu, and Ma}]{Self-Refined}
Song, J.; Zhou, Z.; Liu, J.; Fang, C.; Shu, Z.; and Ma, L. 2023.
\newblock Self-Refined Large Language Model as Automated Reward Function Designer for Deep Reinforcement Learning in Robotics.
\newblock \emph{CoRR}, abs/2309.06687.

\bibitem[{Touvron et~al.(2023)Touvron, Martin, Stone, Albert, Almahairi, Babaei, Bashlykov, Batra, Bhargava, Bhosale, Bikel, Blecher, Canton{-}Ferrer, Chen, Cucurull, Esiobu, Fernandes, Fu, Fu, Fuller, Gao, Goswami, Goyal, Hartshorn, Hosseini, Hou, Inan, Kardas, Kerkez, Khabsa, Kloumann, Korenev, Koura, Lachaux, Lavril, Lee, Liskovich, Lu, Mao, Martinet, Mihaylov, Mishra, Molybog, Nie, Poulton, Reizenstein, Rungta, Saladi, Schelten, Silva, Smith, Subramanian, Tan, Tang, Taylor, Williams, Kuan, Xu, Yan, Zarov, Zhang, Fan, Kambadur, Narang, Rodriguez, Stojnic, Edunov, and Scialom}]{Llama2}
Touvron, H.; Martin, L.; Stone, K.; Albert, P.; Almahairi, A.; Babaei, Y.; Bashlykov, N.; Batra, S.; Bhargava, P.; Bhosale, S.; Bikel, D.; Blecher, L.; Canton{-}Ferrer, C.; Chen, M.; Cucurull, G.; Esiobu, D.; Fernandes, J.; Fu, J.; Fu, W.; Fuller, B.; Gao, C.; Goswami, V.; Goyal, N.; Hartshorn, A.; Hosseini, S.; Hou, R.; Inan, H.; Kardas, M.; Kerkez, V.; Khabsa, M.; Kloumann, I.; Korenev, A.; Koura, P.~S.; Lachaux, M.; Lavril, T.; Lee, J.; Liskovich, D.; Lu, Y.; Mao, Y.; Martinet, X.; Mihaylov, T.; Mishra, P.; Molybog, I.; Nie, Y.; Poulton, A.; Reizenstein, J.; Rungta, R.; Saladi, K.; Schelten, A.; Silva, R.; Smith, E.~M.; Subramanian, R.; Tan, X.~E.; Tang, B.; Taylor, R.; Williams, A.; Kuan, J.~X.; Xu, P.; Yan, Z.; Zarov, I.; Zhang, Y.; Fan, A.; Kambadur, M.; Narang, S.; Rodriguez, A.; Stojnic, R.; Edunov, S.; and Scialom, T. 2023.
\newblock Llama 2: Open Foundation and Fine-Tuned Chat Models.
\newblock \emph{CoRR}, abs/2307.09288.

\bibitem[{Vemprala et~al.(2023)Vemprala, Bonatti, Bucker, and Kapoor}]{ChatGPTforRobotics}
Vemprala, S.; Bonatti, R.; Bucker, A.; and Kapoor, A. 2023.
\newblock ChatGPT for Robotics: Design Principles and Model Abilities.
\newblock \emph{CoRR}, abs/2306.17582.

\bibitem[{Wilson, Fern, and Tadepalli(2012)}]{WilsonFT12}
Wilson, A.; Fern, A.; and Tadepalli, P. 2012.
\newblock A Bayesian Approach for Policy Learning from Trajectory Preference Queries.
\newblock In \emph{Advances in Neural Information Processing Systems 25: 26th Annual Conference on Neural Information Processing Systems 2012. Proceedings of a meeting held December 3-6, 2012, Lake Tahoe, Nevada, United States}, 1142--1150.

\bibitem[{Yarats et~al.(2022)Yarats, Fergus, Lazaric, and Pinto}]{YaratsFLP22DRQV2}
Yarats, D.; Fergus, R.; Lazaric, A.; and Pinto, L. 2022.
\newblock Mastering Visual Continuous Control: Improved Data-Augmented Reinforcement Learning.
\newblock In \emph{The Tenth International Conference on Learning Representations, {ICLR} 2022, Virtual Event, April 25-29, 2022}. OpenReview.net.

\bibitem[{Yin et~al.(2023)Yin, Fu, Zhao, Li, Sun, Xu, and Chen}]{mllmsurvey}
Yin, S.; Fu, C.; Zhao, S.; Li, K.; Sun, X.; Xu, T.; and Chen, E. 2023.
\newblock A Survey on Multimodal Large Language Models.
\newblock \emph{CoRR}, abs/2306.13549.

\bibitem[{Yu et~al.(2019)Yu, Quillen, He, Julian, Hausman, Finn, and Levine}]{metaworld}
Yu, T.; Quillen, D.; He, Z.; Julian, R.; Hausman, K.; Finn, C.; and Levine, S. 2019.
\newblock Meta-World: {A} Benchmark and Evaluation for Multi-Task and Meta Reinforcement Learning.
\newblock In \emph{3rd Annual Conference on Robot Learning, CoRL 2019, Osaka, Japan, October 30 - November 1, 2019, Proceedings}, volume 100 of \emph{Proceedings of Machine Learning Research}, 1094--1100. {PMLR}.

\bibitem[{Yu et~al.(2023)Yu, Gileadi, Fu, Kirmani, Lee, Gonzalez~Arenas, Lewis~Chiang, Erez, Hasenclever, Humplik, Ichter, Xiao, Xu, Zeng, Zhang, Heess, Sadigh, Tan, Tassa, and Xia}]{yu2023language}
Yu, W.; Gileadi, N.; Fu, C.; Kirmani, S.; Lee, K.-H.; Gonzalez~Arenas, M.; Lewis~Chiang, H.-T.; Erez, T.; Hasenclever, L.; Humplik, J.; Ichter, B.; Xiao, T.; Xu, P.; Zeng, A.; Zhang, T.; Heess, N.; Sadigh, D.; Tan, J.; Tassa, Y.; and Xia, F. 2023.
\newblock Language to Rewards for Robotic Skill Synthesis.

\bibitem[{Zeng et~al.(2023)Zeng, Attarian, Ichter, Choromanski, Wong, Welker, Tombari, Purohit, Ryoo, Sindhwani, Lee, Vanhoucke, and Florence}]{SocraticModels}
Zeng, A.; Attarian, M.; Ichter, B.; Choromanski, K.~M.; Wong, A.; Welker, S.; Tombari, F.; Purohit, A.; Ryoo, M.~S.; Sindhwani, V.; Lee, J.; Vanhoucke, V.; and Florence, P. 2023.
\newblock Socratic Models: Composing Zero-Shot Multimodal Reasoning with Language.
\newblock In \emph{The Eleventh International Conference on Learning Representations, {ICLR} 2023, Kigali, Rwanda, May 1-5, 2023}. OpenReview.net.

\bibitem[{Zhang, Li, and Bing(2023)}]{Video-LLaMA}
Zhang, H.; Li, X.; and Bing, L. 2023.
\newblock Video-LLaMA: An Instruction-tuned Audio-Visual Language Model for Video Understanding.
\newblock \emph{CoRR}, abs/2306.02858.

\bibitem[{Zhu et~al.(2023)Zhu, Chen, Shen, Li, and Elhoseiny}]{minigpt4}
Zhu, D.; Chen, J.; Shen, X.; Li, X.; and Elhoseiny, M. 2023.
\newblock MiniGPT-4: Enhancing Vision-Language Understanding with Advanced Large Language Models.
\newblock \emph{CoRR}, abs/2304.10592.

\end{thebibliography}

\end{document}